\documentclass[journal,twoside,web]{ieeecolor}
\usepackage{generic}
\usepackage{cite}
\usepackage{amsmath,amssymb,amsfonts}
\usepackage{algorithmic}
\usepackage{graphicx}
\usepackage{algorithm,algorithmic}
\usepackage{hyperref}
\usepackage{booktabs}
\usepackage{multirow}
\usepackage{makecell}
\hypersetup{hidelinks=true}
\usepackage{textcomp}
\def\BibTeX{{\rm B\kern-.05em{\sc i\kern-.025em b}\kern-.08em
    T\kern-.1667em\lower.7ex\hbox{E}\kern-.125emX}}
\begin{document}
\title{AG-TAL: Anatomically-Guided Topology-Aware Loss for Multiclass Segmentation of the Circle of Willis Using Large-Scale Multi-Center Datasets}

\author{Jialu Liu, Yue Cui, and Shan Yu
	\thanks{This study was supported by the National Key R$ \& $D Program of China (Grant No. 2025YFE0151200), STI 2030- Major Projects (No. 2021ZD0200402), the National Natural Science Foundation of China (No. 82371486), and the Beijing Natural Science Foundation Haidian Original Innovation Joint Fund Project (L232035, L222154). (Corresponding author: Yue Cui and Shan Yu)}
	\thanks{Jialu Liu is with the Laboratory of Brain Atlas and Brain-inspired Intelligence, Institute of Automation, Chinese Academy of Sciences, Beijing, China, also with State Key Laboratory of Brain Cognition and Brain-inspired Intelligence Technology, Chinese Academy of Sciences, Beijing, China, and also with School of Artificial Intelligence, University of Chinese Academy of Sciences, Beijing, China (e-mail: liujialu2023@ia.ac.cn)}
	\thanks{Yue Cui and Shan Yu are with the Laboratory of Brain Atlas and Brain-inspired Intelligence, Institute of Automation, Chinese Academy of Sciences, Beijing, China, also with State Key Laboratory of Brain Cognition and Brain-inspired Intelligence Technology, Chinese Academy of Sciences, Beijing, China, also with School of Artificial Intelligence, University of Chinese Academy of Sciences, Beijing, China, and also with School of Future Technology, University of Chinese Academy of Sciences, Beijing, China (e-mail: yue.cui@nlpr.ia.ac.cn; shan.yu@nlpr.ia.ac.cn)}}

\maketitle

\begin{abstract}
	Accurate multiclass segmentation of the Circle of Willis (CoW) is essential for neurovascular disease management but remains challenging due to complex vascular topology and variable morphology. Existing deep learning methods often suffer from vascular discontinuities and inter-class misclassification, while current topological loss functions incur prohibitive computational costs in 3D multiclass settings. To address these limitations, we propose an Anatomically-Guided Topology-Aware Loss (AG-TAL) and introduce a large-scale, multi-center CoW dataset with unified annotations to facilitate robust model training. AG-TAL specifically integrates a radius-aware Dice loss to address class imbalance in small vessels, a breakage-aware clDice loss that utilizes group convolutions to efficiently preserve local connectivity, and an adjacency-aware co-occurrence loss that leverages anatomical priors to enforce distinct boundaries between neighboring arteries. Evaluated using 5-fold cross-validation, AG-TAL achieved an average Dice score of 80.85\% for all CoW arteries, with small arteries notably higher by 1.05-3.09\% compared to state-of-the-art methods. Across six independent datasets, the performance of AG-TAL achieved Dice scores ranging from 74.46\% to 81.17\% for all CoW arteries, with improvements of 2.20\% to 9.98\% for small arteries compared to other methods. This study demonstrates the superiority of AG-TAL in identifying multiclass CoW arteries and its ability to generalize well to multiple independent datasets. Furthermore, reliability analyses and clinical applications in an Alzheimer’s disease cohort validate the AG-TAL's robustness and its potential for discovering imaging-based morphological biomarkers.
	
\end{abstract}


\section{Introduction}
\label{sec:introduction}

\IEEEPARstart{T}{he} brain receives its blood supply from bilateral internal carotid and vertebral arteries. These vessels anastomose at the base of the brain to form the Circle of Willis (CoW), a polygonal network of arteries that differ in diameter, length, and morphology. As the principal collateral pathway, the CoW delivers blood to the anterior, middle, and posterior cerebral territories as well as the cerebellum. Anatomical alterations of the CoW such as stenosis, occlusion, dilation or branch extension may increase the risks of neurovascular and neurodegenerative disorders, including stroke \cite{b1} and dementia \cite{b2,b3}. Therefore, accurate labeling and quantification of vascular segments are crucial for monitoring and managing brain health. However, manual CoW labeling is labor-intensive and subjective. 

Deep learning techniques with U-shaped architectures offer the potential for angiography data to be used to automatically capture intrinsic arterial characteristics, allowing for a more accurate identification of multiple arteries \cite{b4,b5,b6}. 

However, vascular discontinuities in small vessels and misclassification between adjacent arteries for multiclass recognition still exist, and these may largely reflect insufficient learning of vascular topology. To address these challenges, some studies have integrated topological constraints directly into the loss by quantifying the discrepancies in connectivity between prediction and ground truth, thereby enabling the loss to explicitly preserve vascular structural continuity during training, with persistent homology (PH)-based \cite{b7,b8,b9,b10,b11} improvements and skeleton-based \cite{b12,b13,b14,b15,b16} approaches. 
PH, an algebraic topology tool, imposes constraints by tracking the evolution of topological features across thresholds \cite{b17}. Although PH-based losses such as HuTopo \cite{b9} and BettiMatching \cite{b7,b10} have shown promising performance in 2D vascular segmentation, they suffer from excessive computational overhead, since persistence matching must run on CPUs even after optimization \cite{b11}, leading to impractically long training times for 3D multiclass segmentation. 

Alternatively, skeleton-based methods enhance vascular segmentation by modeling centerlines to preserve structural continuity, with GPU-friendly derivatives ensuring faster training. For instance, centerline-Dice (clDice) \cite{b12,b13} and centerline boundary Dice (cbDice) \cite{b14} have been proposed to incorporate vascular skeleton into the Dice loss. The clDice loss enforces the overlap between the differential centerlines extracted from prediction and ground truth, ensuring that the predicted vascular structure remains topologically coherent. The cbDice computes the vascular diameter from predictions and ground truths, applying reverse weighting to the centerline for clDice, improving vascular boundary precision while maintaining topology. In contrast, the centerline-Cross Entropy (clCE) \cite{b15} loss integrates centerline constraints into the Cross-Entropy (CE) loss, preserving topological integrity and improving segmentation in terms of ground truth overlap. The computation of differential skeletons for prediction incurs high memory consumption, making it challenging to directly apply to multiclass segmentation tasks (e.g., nearly 40GB for more than 7 classes) \cite{b16}. While SkeletonRecall (SkelRecall) \cite{b16} mitigates this by precomputing skeletons to reduce resource usage, its reliance on a recall-only constraint inherently favors sensitivity at the expense of topological precision, failing to penalize false positives. Furthermore, existing losses typically treat arteries in isolation, ignoring anatomical adjacency priors. Some arteries physically adjacent but distinct in the CoW are highly prone to inter-class misclassification, due to the lack of anatomical adjacency constraints.

To address the challenges of accurate multiclass CoW segmentation, we propose an \textbf{anatomically-guided topology-aware loss (AG-TAL)} and construct a large-scale, precisely annotated CoW dataset to facilitate robust model training and evaluation across diverse imaging conditions. AG-TAL is explicitly designed to overcome the aforementioned limitations of prior methods by integrating complementary anatomical and topological cues in a memory-efficient manner. Specifically, we introduce a breakage-aware clDice loss that utilizes group convolutions to extract local contextual connectivity cues, which are integrated into the binary clDice, maintaining memory efficiency in multiclass settings. To further resolve inter-class misclassification caused by physical proximity, we introduce an adjacency-aware co-occurrence loss, which leverages a predefined anatomical adjacency matrix to enforce distinct boundaries between neighboring arteries. The main contributions of this work can be summarized as follows:

\begin{itemize}
	\item We propose the AG-TAL, which unifies topological consistency, inter-segment adjacency, and vessel radius information to jointly enhance continuity and artery discrimination in multiclass vascular segmentation. It consists of a radius-aware Dice loss, a breakage-aware clDice loss and an adjacency-aware co-occurrence loss.
	\item Unlike standard Dice loss, the \textbf{radius-aware Dice loss} utilizes vascular radius information of ground truth as localized weighting into the Dice loss, so that the model is guided to focus more on small vessel segmentation.
	\item To resolve the sensitivity-precision imbalance found in SkelRecall and the high memory cost of clDice, the \textbf{breakage-aware clDice loss} employs group convolutions to capture local 3D contextual connectivity, selectively penalizing voxels associated with vascular breakage while balancing memory efficiency with topological precision.
	\item The \textbf{adjacency-aware co-occurrence loss} leverages a predefined adjacency matrix as anatomical prior knowledge and uses the co-occurrence features to guide the model to learn correct inter-class relationships between different arteries, significantly reducing the issue of inter-class misclassification in the CoW.
	\item In addition, we construct a multi-center annotated CoW dataset with consistent anatomical labeling criteria, which provides the necessary diversity in inter-subject and inter-scanner variability to benchmark multiclass vascular segmentation reliably, supporting both robust training and reproducible cross-center validation.
\end{itemize}

\section{Materials and methods}

\subsection{MRA datasets}
A total of 1341 human time-of-flight magnetic resonance angiography (TOF-MRA) images from 14 publicly available scanning centers with diverse acquisition parameters were used to develop and validate AG-TAL (see Tab.~\ref{tab_1} for subject details and acquisition parameters). The raw data were taken from the publicly available ADAM \cite{b24}, BraVa \cite{b18}, CHUV \cite{b19}, ICBM \cite{b20} (with MNI, UTHC and UCLA sites), IXI \cite{b21} (with Guys, HH and IOP sites), MIDAS \cite{b22}, OASIS \cite{b23} and TopCoW \cite{b25} datasets. 
The OASIS dataset comprises both cognitively normal subjects (randomly divided into OASIS-Normal1 and OASIS-Normal2) and Alzheimer's Disease (AD) patients (OASIS-AD). For test–retest reliability, the Midnight Scan Club (MSC)~\cite{b29} dataset, comprising four repeated scans from 10 healthy subjects, was included without requiring manual annotation.
This study was approved by the Institutional Review Board/Ethic Committee of Chinese Academy of Sciences Institute of Automation.

The CoW comprises bilateral internal carotid arteries (ICA), anterior cerebral arteries (ACA), middle cerebral arteries (MCA), posterior communicating arteries (Pcom), posterior cerebral arteries (PCA), anterior communicating artery (Acom), basilar artery (BA), anterior choroidal arteries (AChA) and superior cerebellar arteries (SCA). The ACA is divided into two parts: ACA1 extends from the ICA bifurcation to the Acom, while ACA2 extends from the Acom to the first branching point. Similarly, the PCA is divided into two parts: PCA1 originates from the BA and extends to the Pcom, while PCA2 extends from the Pcom to the first bifurcation. Together, these vessels form the 20 arteries of the CoW. These vascular segments were manually annotated across 14 datasets, except TopCoW, which has release 12 out of 20 CoW arteries.

% in the first row
\begin{table}[]
	\caption{Subject demographics and TOF-MRA acquisition parameters.}
	\label{tab_1}
	\setlength{\tabcolsep}{3pt}
	\begin{tabular}{@{}p{55pt}p{10pt}p{36pt}p{30pt}p{44pt}p{52pt}@{}}
		\toprule
		\textbf{Datasets} & \textbf{N} & \textbf{Age (SD)} & \textbf{Gender (F/M)} &  \textbf{Scanner} & \textbf{Resolution ($ mm^{3} $)} \\\midrule
		
		\multicolumn{6}{l}{Train and test (n = 926)} \\ \midrule
		BraVa             & 26                & 28.7 (7.1)               & 21 / 5  &3T Siemens &  0.62$\times$0.62$\times$0.62           \\
		CHUV              & 50                & —                             & —  &3T Siemens  \par/ Philips    &  0.41$\times$0.41$\times$0.55      \\
		ICBM-MNI          & 34                & 43.6 (16.5)              & 19 / 15  &1.5T Siemens  &   0.63$\times$0.63$\times$0.60     \\
		IXI-Guys          & 314               & 51.8 (16.2)              & 155 / 159    & 1.5T Philips  &  0.47$\times$0.47$\times$0.80     \\
		IXI-HH            & 181               & 47.4 (16.8)              & 94 / 87      & 3T Philips &  0.47$\times$0.47$\times$0.80      \\
		IXI-IOP           & 73                & 42.5 (16.7)              & 48 / 25      & 1.5T GE  &  0.26$\times$0.26$\times$0.80   \\
		MIDAS             & 109               & 42.6 (14.2)              & 53 / 56    &3T Siemens &   0.51$\times$0.51$\times$0.80       \\
		OASIS-Normal1      & 139               & 64.1 (8.6)              & 78 / 51    &3T Siemens  &0.30$\times$0.30$\times$0.60       \\ \midrule
		\multicolumn{6}{l}{Independent (n = 415)} \\ \midrule           
		ADAM              & 11                & —                             & —       & 1.5T / 3T \par Philips     &0.31$\times$0.31$\times$0.70    \\
		ICBM-UTHC         & 18                & 30.8 (6.4)               & 11 / 7       &3T Siemens  &0.62$\times$0.62$\times$0.62       \\
		ICBM-UCLA         & 130               & 43.8 (15.2)              & 64 / 66     &3T Siemens  &0.62$\times$0.62$\times$0.62        \\
		OASIS-AD          & 49                & 73.0 (7.2)                             & 15 / 34          &3T Siemens &0.30$\times$0.30$\times$0.60    \\
		OASIS-Normal2     & 82                & 66.4 (9.2)                             & 43 / 39      &3T Siemens   &0.30$\times$0.30$\times$0.60       \\
		TopCoW2024        & 125               & —                             & —     &1.5T / 3T \par Siemens   &0.36$\times$0.36$\times$0.50        \\ \midrule
		\multicolumn{6}{l}{Test-retest (n = 10 with 40 scans)} \\ \midrule 
		MSC        & 10               & 29.1 (3.2)                            & 5 / 5     &3T Siemens   &0.63$\times$0.63$\times$1.00        \\
		\bottomrule
	\end{tabular}
\end{table}

\subsection{MRA pre-processing}
All images were first reoriented to the standard RAS (Right–Anterior–Superior) coordinate system to avoid orientation-induced segmentation errors. To focus on the CoW, a specific ROI was extracted via template-to-subject registration and cropping. A publicly available time-of-flight (TOF) atlas in MNI space was used as the template \cite{b4}, and an $ 80 \times80 \times 80 $ voxel cubic ROI at 1 mm isotropic resolution was defined in the same spatial domain, which fully covers the entire the CoW region. Registration was executed via the ANTs \textit{antsRegistrationSyNQuick.sh} utility~\cite{b26}. The registered ROI mask was applied to crop the native images and annotations for subsequent model training and inference.

\subsection{Anatomically-guided topology-aware loss}
The proposed AG-TAL integrating topological consistency, inter-segment adjacency and radius information of the CoW is illustrated in Fig. 1 and described in detail below. AG-TAL consists of radius-aware Dice, breakage-aware clDice and adjacency-aware co-occurrence losses. The total loss is the combination of AG-TAL and the CE loss. The                                                                                                                                                                                                                                                                                                                                                                                                                                                                                                                                                                                                                                                                                                                                                                                                                                                                                                                                                                                                                                                                                                                                                                                                                                                                                                                                                                                                                                                                                                                                                                                                                                                                                                                                                                                                               overall segmentation loss function is shown in Eq. \ref{AG-TAL}, where $ \lambda_1 $ and $ \lambda_2 $ are hyperparameters.
\begin{align}
	L_{AG-TAL}&=L_{CE}+L_{Dice\_Weight}+\lambda_1L_{Neighbors\_clDice} \notag\\
	&+\lambda_2L_{Co-occurence\_Dice}\label{AG-TAL}
\end{align}

We denote $ V_P,V_L\in\mathbb{R}^{C\times H\times W\times D} $ as a segmentation map and a ground truth map respectively, where $ H,W,D $ indicate the height, width and depth of the image, and $ C $ is the number of vascular segments.

\begin{figure*}[]
	%	\columnwidth
	\includegraphics[width=\textwidth]{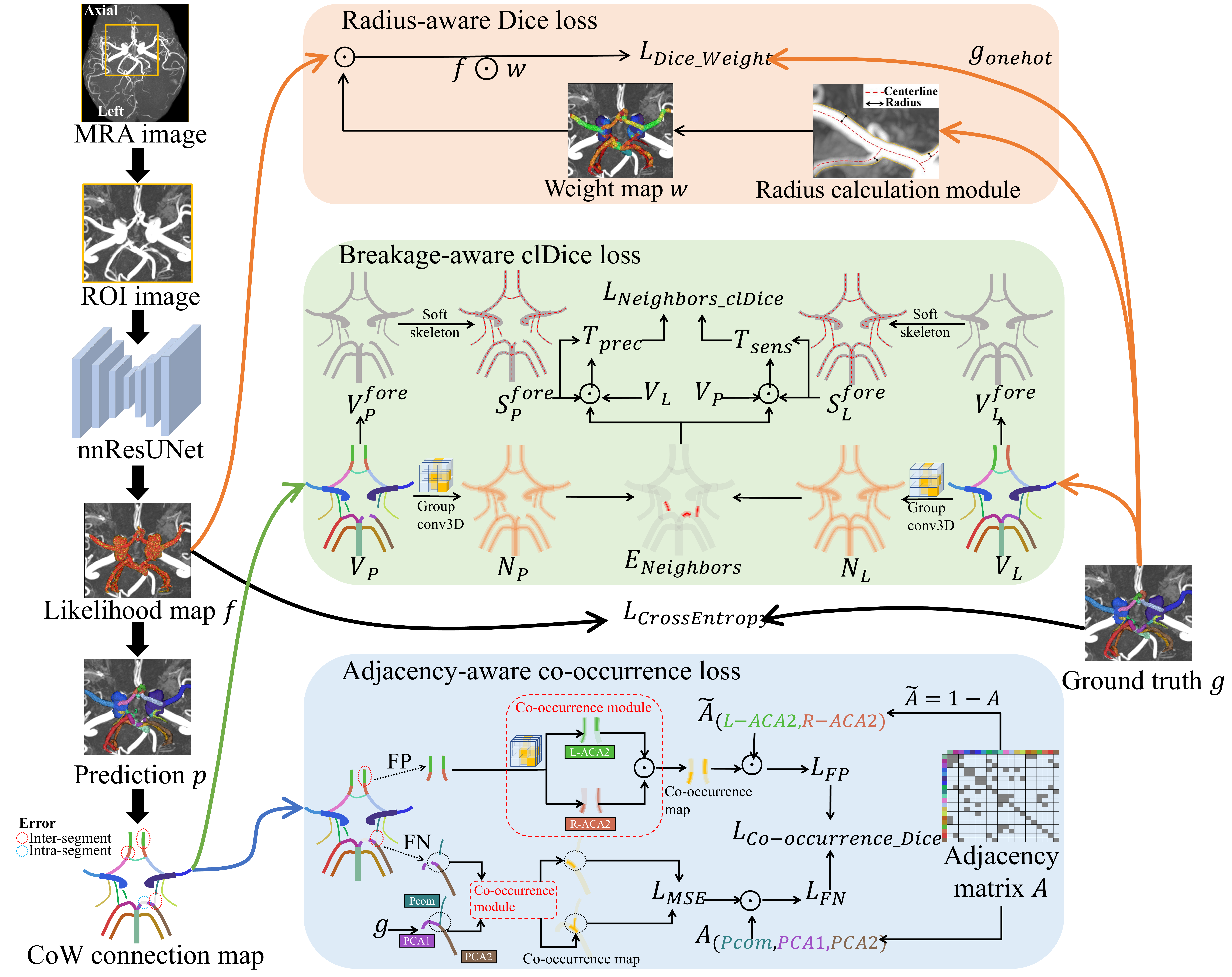}
	\caption{Overview of the proposed AG-TAL. AG-TAL consists of radius-aware Dice, breakage-aware clDice and adjacency-aware co-occurrence loss functions. The total loss is the combination of AG-TAL and the CE loss.}
	\label{fig_1}
\end{figure*}

\subsubsection{Radius-aware Dice loss}
The vascular radius map, $ Radius $, is computed utilizing the ground truth and estimated with the technique proposed by Meng et al.~\cite{b27}. The maximum value $ R_{max}$ and minimum value $R_{min} $ in the $ Radius $ are extracted, and the normalization technique is used to design the weight of voxels as shown in Eq. \ref{radius_weight}, where $ v\in V_L $.
\begin{gather}
	W_{Radius}(v)=
	\begin{cases}
		1 &v\in BG \\
		\exp{\left(\frac{R_{max}-Radius(v)}{R_{max}-R_{min}}\right)} &v\in CoW \label{radius_weight}
	\end{cases}
\end{gather}

Voxels with smaller radii are allocated higher weight, whereas background voxels are uniformly assigned a weight value of 1. 
The modified loss function, shown in Eq.~\ref{dice_weight loss}, is obtained by adding the $ W_{Radius}\in\mathbb{R}^{1\times H\times W\times D} $ to the dice loss.
\begin{align}
	L_{Dice\_Weight} &=-Dice((V_{L},V_{P},W_{Radius}) \notag\\
	&=\frac{-2|(V_{L}\cap V_{P}) \times W_{Radius}|}{|V_{L}\times W_{Radius}|+|V_{P}\times W_{Radius}|} \label{dice_weight loss}
\end{align}

\subsubsection{Breakage-aware clDice loss}
Inspired by the manual annotation process where experts examine local neighborhoods to ensure connectivity, we propose a vessel breakage-aware clDice Loss to enhance feature learning at break-prone locations.
First, neighborhood similarity maps for the prediction and the ground truth are computed using a $  3\times3\times3 $ group convolution with weights fixed to 1. This operation, equivalent to multiplying an $ 3\times3\times3 $ average pooling result by the sum of the kernel weights, sums class-specific probabilities within the local neighborhood. The resulting sum is higher near the vessel centerline and lower at the edges or in the background. From $ V_P$ and $V_L $, the neighborhood similarity error map $ E_{N}\in\mathbb{R}^{C\times H\times W\times D} $ is derived via the L1 distance (Eq. \ref{E_neighbor}).
\begin{align}
	E_{N}=|GroupConv3D(V_P)-GroupConv3D(V_L)|
	\label{E_neighbor}
\end{align}

$ E_{N}$ highlights vessel breakages and false positives while de-emphasizing minor boundary mismatches, aligning with the clinical priority of topological integrity over exact boundary delineation.
Original clDice loss is computationally prohibitive for multiclass tasks due to the memory cost of per-channel skeletonization. Our approach circumvents this by treating all vascular classes as a single CoW foreground entity to extract a unified soft-skeleton. Then the computationally cheaper multiclass error map $ E_{N} $ as a spatial weight. This focuses the model on topologically significant errors across all classes without massive overhead. 

The modified clDice loss is formulated as follows. Here, $ S_P^{fore} $ and $ S_L^{fore}\in\mathbb{R}^{1\times H\times W\times D} $ denote the binary-foreground soft-skeletons, $Tprec$ and $Tsens$ denote the weighted topology precision and sensitivity, and $\odot$ denotes the Hadamard product.
\begin{align}
	&Tprec=\frac{|(S_P^{fore}\odot V_L )\odot E_{N}|}{|S_P^{fore}\odot E_{N}|} \notag\\
	&Tsens=\frac{|(S_L^{fore}\odot V_P )\odot E_{N}|}{|S_L^{fore}\odot E_{N}|} \notag\\
	&L_{Neighbors\_clDice}=-2\frac{Tsens \times Tprec}{Tsens+Tprec}
	\label{Neighbors_cldice}
\end{align}

\subsubsection{Adjacency-aware co-occurrence loss}
The adjacency-aware co-occurrence loss leverages an anatomical adjacency matrix to guide the network in learning the correct inter-class relationships and reducing misclassifications. 
It addresses two failure modes: incorrect inter-class connections (false positives) and missing expected connections (false negatives).

Consequently, the adjacency-aware co-occurrence loss is designed based on the ideas of ``penalizing incorrect adjacency relationships" and ``promoting correct adjacency relationships" respectively. Based on anatomical priors, we define an adjacency matrix $ A\in \{0,1\}^{C\times C} $ (Eq. \ref{Aij}), where $ A_{ij}=1 $ if artery $ i $ and $ j $ are connected, and $ A_{ij}=0 $ otherwise.
\begin{gather}
	A_{ij}=
	\begin{cases}
		1 & (i,j)\ \text{is an adjacent pair} \\
		0 &(i,j)\ \text{is a non-adjacent pair} 
	\end{cases} \label{Aij}
\end{gather}

\paragraph{\textbf{Penalizing incorrect adjacency relationships}} Following the design strategy of co-occurrence features proposed by Zhang et al.~\cite{b28} for different objects, the class co-occurrence relations in the multiclass segmentation task of the CoW were defined based on an anatomical prior adjacency matrix. Let $ V_P^i$ be the predicted probability map for artery $ i $. For a voxel $ v\in V_P $, the local occurrence probability $ P_i(v)$ is estimated via by aggregating its predicted probabilities within neighborhood: $ P_i(v)=V_P^i(v)+\sum_{n\in N(v)}{V_P^i(n)} $, where $ N(v) $ represents a $ 3\times3\times3 $ neighborhood. The joint occurrence probability of classes $i$ and $ j $ is defined as $ p_{ij}(v)=\ P_i(v)P_j(v) $. For an anatomically non-adjacent pair $ \left(i,j\right) $, the constraint $ P_i(v)P_j(v)=0 $ should be enforced.

To apply this penalty across the entire image, the co-occurrence degree $ C_{ij} $ of the pair $ \left(i,j\right) $ is defined as Eq. \ref{Cij}:
\begin{equation}
	C_{ij}=\sum_{v\in V_P}p_{ij}(v)=\sum_{v\in V_P}P_i(v)P_j(v)
	\label{Cij}
\end{equation}

Since the objective is to ensure that the co-occurrence degree of all non-co-occurrence pairs approaches zero in the prediction, the adjacency matrix $ A $ is  inverted to obtain the non-adjacency matrix $ \widetilde{A}=1-A $.
The overall penalty for false positive connections is then computed as Eq. \ref{LFP}.
\begin{equation}
	L_{FP}=\frac{1}{|\widetilde{A}|}\sum_{ij}\widetilde{A}_{ij}C_{ij}
	\label{LFP}
\end{equation}

\paragraph{\textbf{Encouraging correct adjacency relationships}} 
Explicitly encouraging the preservation of legitimate vascular inter-class connections in the predictions can ensure topological correctness.
A keypoint region mask $ M_{keypoint} \in \{0,1\}^{1\times H\times W\times D} $ was first extracted from the ground truth to localize connectivity-critical regions. Within the keypoint regions, the joint co-occurrence probability $ p_{ij}(v) $ for each co-occurrence pair $ \left(i,j\right) $ was computed voxel-wise in both the prediction $V_P$ and the ground truth $V_L$. Here, $(\cdot) \in \{pred, gt\}$, $V_{pred}^i = V_P^i$, and $V_{gt}^i = V_L^i$.

To further ensure spatial consistency of inter-class connections, edge information is incorporated using 3D Sobel operators applied via grouped convolution. This yields class-specific edge maps $ {Edge}_i^{(\cdot)} $ that characterize the spatial location of vascular boundaries for each class, where $ K_X,K_Y,K_Z $ denotes Sobel kernels along the x-, y-, and z-axes. 
\begin{align}
	Edge_i^{(\cdot)}=\sqrt{\sum_{d\in \{X,Y,Z\}}\left(K_dV_{(\cdot)}^i\right)^2}\label{edge}
\end{align}

By combining joint co-occurrence strength with edge responses, the loss $L_{FN}$ simultaneously enforcing both the existence and spatial alignment of inter-class connections is formulated, as shown in Eq. \ref{LFN}. Here, a connectivity feature is defined as $q_{ij}^{(\cdot)}(v) = p_{ij}^{(\cdot)}(v) Edge_i^{(\cdot)}(v) $.
\begin{align}
	L_{FN}=\sum_{i,j}A_{ij}\sum_{v\in M_{keypoint}} \left(q_{ij}^{pred}(v)-q_{ij}^{gt}(v) \right)^2
	\label{LFN}
\end{align}

The adjacency-aware co-occurrence Loss is the sum of these two components, jointly promoting anatomically consistent and topologically correct CoW segmentation: 
\begin{equation}
	L_{Co-occurence\_Dice}=L_{FP}+L_{FN}
\end{equation}

\subsection{Experimental Setup}
\subsubsection{Dataset allocation}
Out of the 1341 TOF-MRA images containing manually annotations, 926 cases were used as the training and test set for model development and evaluation via 5-fold cross-validation, with each fold split into training (60\%), validation (20\%), and test (20\%) sets, ensuring that each case served as a test sample once. The remaining 415 cases, drawn from the ADAM, ICBM-UTHC, ICBM-UCLA, OASIS-AD, OASIS-Normal2, and TopCoW datasets, were treated as independent datasets and were not involved in the cross-validation process. For these independent datasets, final segmentation results were obtained by averaging predictions from the five cross-validation models.
\subsubsection{Quantitative Evaluation}

Softmax probability maps were converted to discrete labels by voxel-wise maximization, resulting in 21 classes (20 CoW arteries and background). Quantitative performance was evaluated using Dice, clDice, 95th percentile hausdorff distance (HD95), 0-th Betti number ($ \beta_0 $) error, and Betti number ($ \beta $) error. Specifically, clDice assesses topological connectivity via centerline precision and recall, while HD95 measures boundary error with reduced outlier sensitivity. Topological integrity was quantified through $\beta_0$ error and $\beta$ error (the sum of $\beta_0$ and $\beta_1$ errors, capturing discrepancies in connected components and holes). Metrics were computed per artery and aggregated by vessel radius (large, medium, small) and overall.

\subsubsection{Comparison Experiments and Training Details}
AG-TAL was compared with four state-of-the-art (SOTA) centerline-based losses (cbDice~\cite{b14}, clCE~\cite{b15}, clDice~\cite{b12}, and SkelRecall~\cite{b16}) and two PH-based losses (BettiMatching~\cite{b7} and HuTopo~\cite{b9}). As several methods were not originally designed for multiclass segmentation, appropriate adaptations were applied. For cbDice, clCE, and clDice loss, which rely on differentiable skeletonization, a unified vascular skeleton was extracted for all arteries and multiplied with probability maps or one-hot labels. For PH-based losses, a practical 3D-to-2D strategy was adopted, where predictions and ground truths were projected along the x, y, and z axes using maximum intensity projection (MIP), and PH losses were computed on the resulting 2D images.

The nnResUNet~\cite{b6} served as the backbone, with the baseline loss combining Dice and CE loss. SOTA losses were incorporated into the addition as defined in Eq.~\ref{total_loss}.
\begin{equation}
	L_{total}=L_{baseline}+L_{addition}
	\label{total_loss}
\end{equation}

Models were trained for 200 epochs, using random patches of size $ 160 \times 128 \times 56 $ , with images resampled to a target spacing of $ 0.47 \times 0.47 \times 0.80\ mm^3 $. Optimization was performed using SGD with Nesterov momentum ($ \mu =  $ 0.99) and an initial learning rate of 0.01. 
All models were implemented in PyTorch and trained on a single NVIDIA A40 GPU.

\subsubsection{Ablation study}
\paragraph{\textbf{Model architecture-agnostic experiments}}
To evaluate the architecture-agnostic property of the proposed AG-TAL, experiments were conducted using different backbone architectures. Specifically, the ResUNet backbone was replaced with the transformer-based SwinUNETR~\cite{b33} and the mamba-based U-Mamba~\cite{b34}. For each architecture, models were trained using the baseline loss and using AG-TAL as an additional loss term, while keeping all other training settings unchanged.
\paragraph{\textbf{Impacts of each component}}
To analyze the contribution of each component in AG-TAL, five loss configurations were evaluated: (1) the conventional Dice + CE loss $ L_{Baseline} $; (2) replacing the Dice loss with $ L_{Dice\_Weight} $; (3) combining $ { L}_{Dice\_Weight} $ and $ L_{Neighbors\_clDice} $; (4) combining $ L_{Dice\_Weight} $ and $ L_{Co-occurrence\_Dice} $; and (5) the AG-TAL (proposed). To determine the optimal hyperparameters $ \lambda_1 $ and $ \lambda_2 $, a systematic grid search over the range $[0, 1]$ with step 0.25 identified optimal values $\lambda_1=$ 0.5 and $\lambda_2=$ 1.0, which were used in all experiments.

\paragraph{\textbf{Interpretability analysis}}
For interpretability analysis, Grad-CAM~\cite{b32} was employed to visualize the attention of models trained with different loss configurations. Attention maps were generated for representative arteries to facilitate qualitative comparison across different AG-TAL components.

\subsubsection{Reliability and robustness analysis}
To assess the reliability of AG-TAL, signal-to-noise ratio (SNR) and test–retest analyses were conducted. Correlations between image SNR and Dice scores were analyzed to evaluate robustness against SNR variations. Additionally, test–retest reliability experiments were conducted using the MSC dataset. Mean arterial diameters were computed per scan, and reliability was quantified using the intra-class correlation coefficient (ICC(3,1))~\cite{b30}, with t-SNE used for visualization. 

\subsubsection{AG-TAL application to AD analysis}
To demonstrate the efficacy of AG-TAL on pathological data, a comparative experiment was conducted between cognitively normal controls (NC) and AD patients in the OASIS dataset. Dice score distributions were first compared to ensure comparable segmentation accuracy between cohorts. 
Arterial diameters were then derived from the segmentations and compared across groups, using violin plots for visualization, with significance assessed via a 10000-iteration permutation t-test.
Furthermore, inter-cohort differences across all arteries were analyzed, and correlations between every artery's diameter and clinical cognitive assessment scores of AD cohort were evaluated.

\section{Results}
\subsection{Qualitative and Quantitative Assessment}
\begin{figure}
	\includegraphics[width=\columnwidth]{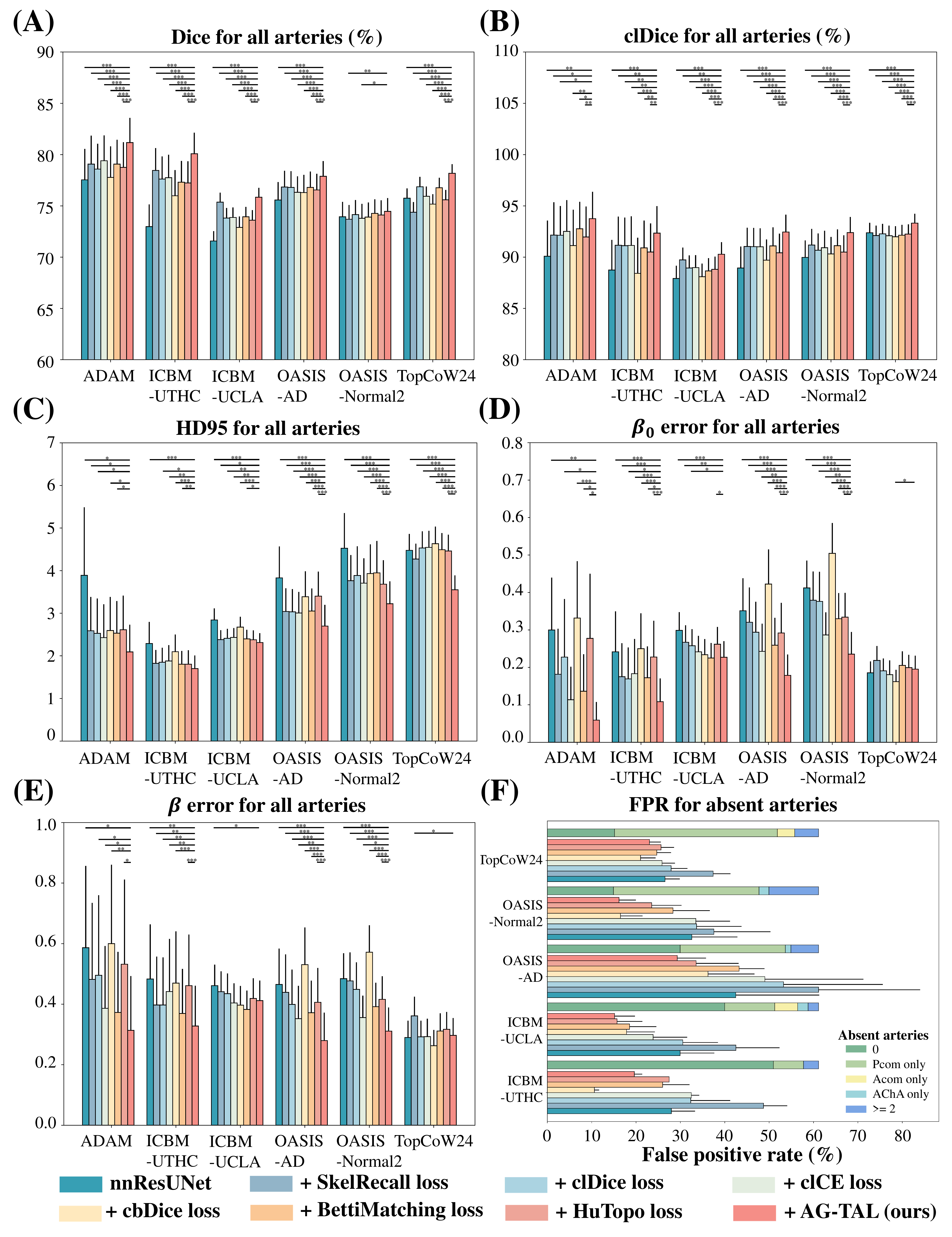}
	
	\caption{Comprehensive comparison of results on independent datasets. \textbf{(A-E).} Quantitative results from eight different methods for six independent datasets were evaluated in terms of Dice, clDice, HD95, $ \beta_0 $ error and $ \beta $ error metrics. $ *, p < 0.05; **, p < 0.01; ***, p < 0.001 $; n.s., not significant using two-sided paired t test with FDR correction for multiple comparisons. \textbf{(F).} Quantitative results of FPR for the absence of small arteries in five independent datasets.}
	\label{fig_bar}
\end{figure}

The comprehensive quantitative results for each method on the test set are presented in Tab. \ref{tab_3}. Dice score, a standard metric for assessing spatial overlap in vessel segmentation \cite{b4,b6,bffcm}, was used as the primary performance indicator. Specifically, on the test set, AG-TAL surpassed all competing methods across large, medium, and small arteries, achieving improvements of 0.33\%, 0.91\%, 1.82\% and 1.05\% over the baseline, 0.38\%, 0.81\%, 1.05\% and 0.80\% over SkelRecall, and 0.34\%, 0.77\%, 1.26\% and 0.82\% over BettiMatching. Notably, AG-TAL consistently ranked first in topology-aware ($\text{clDice}$, $\beta_0$, and $\beta$ error) and distance-based (HD95) metrics. On the independent datasets (Fig. \ref{fig_bar}), AG-TAL demonstrated robust generalization, with average Dice improvements ranging from 2.53\% to 6.51\% for medium, 2.20\% to 9.98\% for small, and 0.53\% to 7.10\% for all vessels overall compared to other methods. These results confirm that AG-TAL effectively preserves fine-scale vascular topology and maintains stable accuracy across diverse imaging domains and acquisition sites.

False positive rate (FPR) analysis for anatomically absent arteries is shown in Fig. \ref{fig_bar}F. Substantial inter-dataset variability was observed, particularly in the OASIS-Normal2 cohort (24.4\% variation-free) and the TopCoW24 dataset (24.8\% variation-free). AG-TAL achieved the lowest FPR in three of the five independent datasets, and the second lowest FPR in two datasets, indicating its ability to avoid erroneous segmentation of congenitally absent arteries.

\begin{table}[]
	\caption{Comparative experimental results on the test datasets for large (ICA and BA),medium (PCA, MCA, ACA, and SCA),small (Pcom, Acom, and AChA) and all arteries.}
	\label{tab_3}
	\setlength{\tabcolsep}{3pt}
	%	Metrics
	\begin{tabular}{p{59pt}>{\centering\arraybackslash}p{41pt}>{\centering\arraybackslash}p{41pt}>{\centering\arraybackslash}p{41pt}>{\centering\arraybackslash}p{41pt}}
		\toprule
		\multicolumn{1}{c}{\textbf{Methods}} & \multicolumn{1}{c}{\textbf{Large}}       & \multicolumn{1}{c}{\textbf{Medium}}      & \multicolumn{1}{c}{\textbf{Small}}       & \multicolumn{1}{c}{\textbf{All}}         \\ \midrule
		\multicolumn{5}{l}{\textbf{Dice(\%)$ \uparrow $}}   \\ \midrule
		nnResUNet                            & 89.42±4.52           & 81.07±8.60           & 70.98±16.16          & 79.80±9.87           \\
		+ SkelRecall                         & 89.37±4.39           & 81.17±8.64           & 71.75±15.90          & 80.05±9.82           \\
		+ clDice                             & 89.27±4.31           & 81.03±8.64           & 71.07±16.05          & 79.77±9.85           \\
		+ clCE                               & 89.46±4.46           & 81.11±8.68           & 71.21±15.99          & 79.89±9.88           \\
		+ cbDice                             & 89.40±4.35           & 80.78±8.85           & 69.71±17.42          & 79.30±10.32          \\
		+ BettiMatching                      & 89.41±4.35           & 81.21±8.49           & 71.54±15.73          & 80.03±9.68           \\
		+ HuTopo                             & 89.13±4.23           & 80.75±8.82           & 70.80±16.29          & 79.52±10.00          \\
		+ AG-TAL (ours)                     & \textbf{89.75±4.09}  & \textbf{81.98±8.17}  & \textbf{72.80±15.34} & \textbf{80.85±9.35}  \\ \midrule
		\multicolumn{5}{l}{\textbf{clDice(\%)$ \uparrow $}} \\ \midrule
		nnResUNet                            & 97.79±4.72           & 94.34±9.39           & 85.82±20.02          & 92.73±11.35          \\
		+ SkelRecall                         & 97.26±5.60           & 94.14±9.85           & 86.14±20.15          & 92.61±11.79          \\
		+ clDice                             & 97.21±5.93           & 94.11±9.76           & 85.87±19.88          & 92.51±11.71          \\
		+ clCE                               & 97.82±4.88           & 94.27±9.59           & 85.93±19.89          & 92.72±11.46          \\
		+ cbDice                             & 97.67±5.04           & 93.84±10.02          & 84.58±21.37          & 92.10±12.11          \\
		+ BettiMatching                      & 97.62±5.23           & 94.38±9.34           & 86.18±19.60          & 92.82±11.29          \\
		+ HuTopo                             & 97.53±5.17           & 93.90±9.80           & 85.69±20.28          & 92.39±11.73          \\
		+ AG-TAL (ours)                     & \textbf{97.95±4.20}  & \textbf{94.83±9.08}  & \textbf{87.45±18.81} & \textbf{93.46±10.78} \\ \midrule
		\multicolumn{5}{l}{\textbf{HD95$ \downarrow $}} \\ \midrule   
		nnResUNet                            & 1.50±3.68          & 2.33±4.47          & 2.71±5.13          & 2.30±4.52          \\
		+ SkelRecall                         & 1.51±3.68          & 2.37±3.92          & 2.75±5.70          & 2.34±4.33          \\
		+ clDice                             & 1.47±3.95          & 2.38±4.43          & 2.70±4.78          & 2.32±4.45          \\
		+ clCE                               & 1.50±4.12          & 2.32±3.87         & 2.68±4.93          & 2.29±4.17          \\
		+ cbDice                             & \textbf{1.36±1.78} & 2.39±3.78          & 2.92±5.14          & 2.37±3.82          \\
		+ BettiMatching                      & 1.47±3.44          & 2.31±3.66          & 2.60±4.48          & 2.26±3.83          \\
		+ HuTopo                             & 1.45±2.86          & 2.47±4.24          & 2.77±4.97          & 2.39±4.22          \\
		+ AG-TAL (ours)                      & 1.48±3.60          & \textbf{2.14±3.82} & \textbf{2.46±4.50} & \textbf{2.12±3.96} \\ \midrule
		\multicolumn{5}{l}{\textbf{$ \beta_0 $ error$ \downarrow $}} \\ \midrule
		nnResUNet                            & 0.08±0.35          & 0.18±0.46          & 0.34±0.66          & 0.21±0.49          \\
		+ SkelRecall                         & 0.13±0.44          & 0.21±0.51          & 0.38±0.70          & 0.24±0.54          \\
		+ clDice                             & 0.23±0.57          & 0.20±0.49          & 0.35±0.67          & 0.24±0.55          \\
		+ clCE                               & 0.07±0.32          & 0.17±0.43          & 0.35±0.68          & 0.20±0.48          \\
		+ cbDice                             & 0.15±0.46          & 0.27±0.55          & 0.35±0.64          & 0.27±0.56          \\
		+ BettiMatching                      & 0.09±0.32          & 0.18±0.44          & 0.34±0.65          & 0.20±0.48          \\
		+ HuTopo                             & 0.13±0.46          & 0.22±0.51          & 0.36±0.69          & 0.24±0.55          \\
		+ AG-TAL (ours)                      & \textbf{0.07±0.29} & \textbf{0.16±0.43} & \textbf{0.29±0.60} & \textbf{0.18±0.45} \\ \midrule
		\multicolumn{5}{l}{\textbf{$ \beta $ error$ \downarrow $}}    \\ \midrule
		nnResUNet                            & 0.47±1.08          & 0.30±0.70          & 0.36±0.68          & 0.34±0.75          \\
		+ SkelRecall                         & 0.63±1.16          & 0.34±0.76          & 0.39±0.72          & 0.39±0.81          \\
		+ clDice                             & 0.67±1.21          & 0.32±0.73          & 0.36±0.68          & 0.38±0.79          \\
		+ clCE                               & \textbf{0.44±0.98} & 0.29±0.68          & 0.36±0.70          & 0.33±0.73         \\
		+ cbDice                             & 0.53±1.08          & 0.39±0.79          & 0.36±0.65          & 0.40±0.80          \\
		+ BettiMatching                      & 0.47±1.03          & 0.30±0.68          & 0.35±0.67          & 0.34±0.73          \\
		+ HuTopo                             & 0.53±1.14          & 0.35±0.75          & 0.38±0.71          & 0.38±0.80          \\
		+ AG-TAL (ours)                     & 0.44±1.01          & \textbf{0.27±0.68} & \textbf{0.30±0.62} & \textbf{0.31±0.72} \\ \bottomrule
	\end{tabular}
\end{table}

\begin{figure*}[]
	\begingroup
	
	\includegraphics[width=\textwidth]{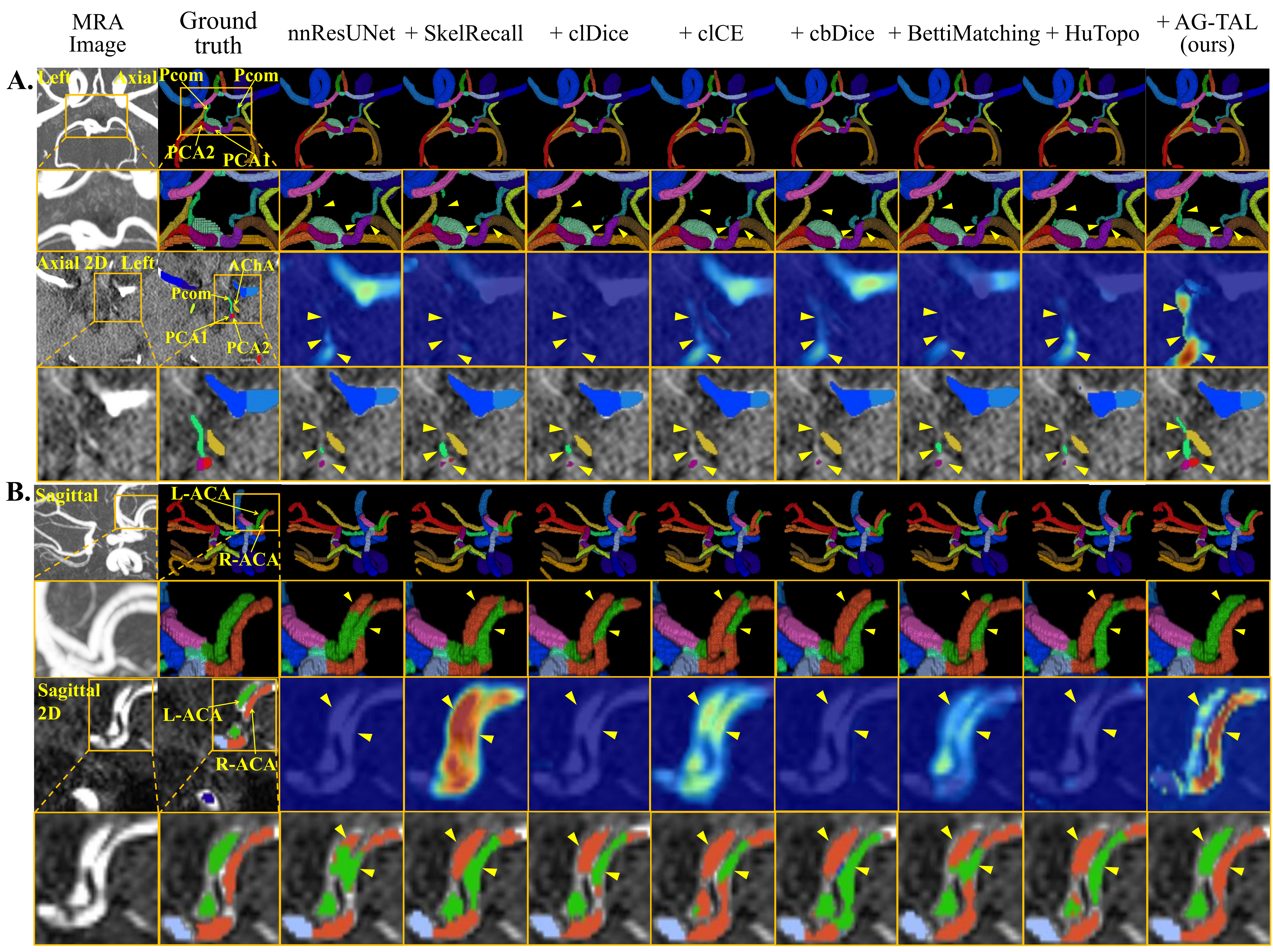}
	\caption{Qualitative results and corresponding attention visualizations for multiclass CoW segmentation from different datasets (OASIS-Normal1, A; IXI-Guys, B). For each subject, four rows are shown: the CoW MIPs of different methods, zoomed-in views of vascular regions, Grad-CAM attention maps on representative 2D views, and the corresponding segmentations. Yellow arrows indicate vessels prone to breakage or misclassification. \textbf{(A).} AG-TAL preserves the continuity of small and low-contrast vessels such as the Pcom and AChA, while accurately localizing the interfaces among Pcom, PCA1, and PCA2. The attention maps show enhanced and continuous responses along fragile vascular regions, supporting improved topological continuity. \textbf{(B).} AG-TAL effectively prevents misclassification between bilateral ACA2, by embedding anatomical connectivity priors. Distinct and asymmetric attention patterns are observed for bilateral ACA2, facilitating discriminative feature learning.}
	\label{fig_3}
	\endgroup
\end{figure*}

Fig. \ref{fig_3} presents a qualitative and interpretability analysis on the test datasets. As shown in Fig. \ref{fig_3}A, AG-TAL demonstrates superior topological integrity, particularly in low-contrast vessels like the Pcom and AChA, which are prone to discontinuities in other methods. Grad-CAM visualizations reveal a continuous, linear attention pattern along these arteries, particularly in low-contrast junctions between the left Pcom and PCA. This indicates that gradient responses are consistently guided by vascular morphology, reflecting the synergy between radius-aware Dice and breakage-aware clDice losses in enforcing neighborhood consistency. Furthermore, Fig. \ref{fig_3}B highlights AG-TAL’s ability to distinguish physically adjacent, confusable arteries (e.g., bilateral ACA2). Unlike other methods, AG-TAL assigns distinct, asymmetric attention to these arteries, facilitating class-specific feature learning. These findings provide interpretable evidence of AG-TAL’s effectiveness in enhancing topological continuity and inter-segment discrimination.

\subsection{Ablation Study}
\subsubsection{Model Architecture-Agnostic Experiments}

As shown in Tab.~\ref{tab_4}, AG-TAL consistently improved segmentation performance across different backbone architectures. With SwinUNETR, AG-TAL increased the Dice score by 1.20\% for small arteries and by 0.68\% for all arteries. Similarly, improvements of 1.33\% and 0.47\% were observed for small and all arteries, respectively, when using U-Mamba. These results indicate that AG-TAL provides robust performance gains across diverse network architectures.

\begin{table}[]
	
	\caption{Quantitative results of Dice score for the ablation study. ($ L_D=L_{Dice\_Weight} $, $ L_N=L_{Neighbors\_clDice} $, $ L_C=L_{Co-occurrence\_Dice} $)}
	\setlength{\tabcolsep}{3pt}
	\begin{tabular}{p{59pt}>{\centering\arraybackslash}p{41pt}>{\centering\arraybackslash}p{41pt}>{\centering\arraybackslash}p{41pt}>{\centering\arraybackslash}p{41pt}}
		%		\multicolumn{1}{c|}{\textbf{Metrics}}
		%	\begin{tabular}{@{}l|lllll@{}}
			\toprule
			\multicolumn{1}{c}{\textbf{Methods}} & \multicolumn{1}{c}{\textbf{large}} & \multicolumn{1}{c}{\textbf{medium}} & \multicolumn{1}{c}{\textbf{small}} & \multicolumn{1}{c}{\textbf{all}} \\ \midrule  
			\multicolumn{5}{l}{Model architecture-agnostic experiments}\\ \midrule
			U-Mamba                              & \textbf{89.51±4.56}                & 81.35±8.72                          & 71.28±16.06                        & 80.06±9.93                       \\
			+ AG-TAL (ours)                             & 89.50±4.54                         & \textbf{81.59±8.33}                 & \textbf{72.61±15.14}               & \textbf{80.53±9.46}              \\ [-1mm] \midrule 
			SwinUNETR                            & 89.01±4.45                         & 79.81±9.61                          & 68.88±17.68                        & 78.46±10.85                      \\
			+ AG-TAL (ours)                             & \textbf{89.19±4.50}                & \textbf{80.40±9.04}                 & \textbf{70.08±16.75}               & \textbf{79.14±10.29}             \\ \midrule
			\multicolumn{5}{l}{Impacts of each component}\\ \midrule
			nnResUNet                            & 89.42±4.52                         & 81.07±8.60                          & 70.98±16.16                        & 79.80±9.87                       \\
			+ $ L_D $                               & 89.21±4.27                         & 80.89±8.00                          & 71.63±14.26                        & 79.82±9.01                       \\
			+ $ L_D+ L_N $                         & 89.21±4.43                         & 81.34±7.94                          & 72.74±13.80                        & 80.37±8.88                       \\
			+ $ L_D+ L_C $                         & 89.56±4.44                         & 81.53±8.17                          & 72.54±14.73                        & 80.49±9.25                       \\
			+ AG-TAL (ours)                             & \textbf{89.75±4.09}                & \textbf{81.98±8.17}                 & \textbf{72.80±15.34}               & \textbf{80.85±9.35}              \\ 
			\bottomrule
		\end{tabular}
		\label{tab_4}
	\end{table}

	\subsubsection{Impacts of Each Component}

	Quantitative results in Tab.~\ref{tab_4} demonstrate that each component of AG-TAL contributes to segmentation performance. $ L_{Dice\_Weight} $ notably improved small vessel segmentation, yielding higher Dice scores. Incorporating $ L_{Neighbors\_clDice} $ further enhanced the continuity of small and medium arteries. $ L_{Co-occurrence\_Dice} $ effectively reduced artery misclassification and improved overall consistency. The full AG-TAL, which integrates all components, achieved the best overall segmentation performance.

	\subsubsection{Interpretability analysis of each AG-TAL component}
	
\begin{figure*}
	\includegraphics[width=\textwidth]{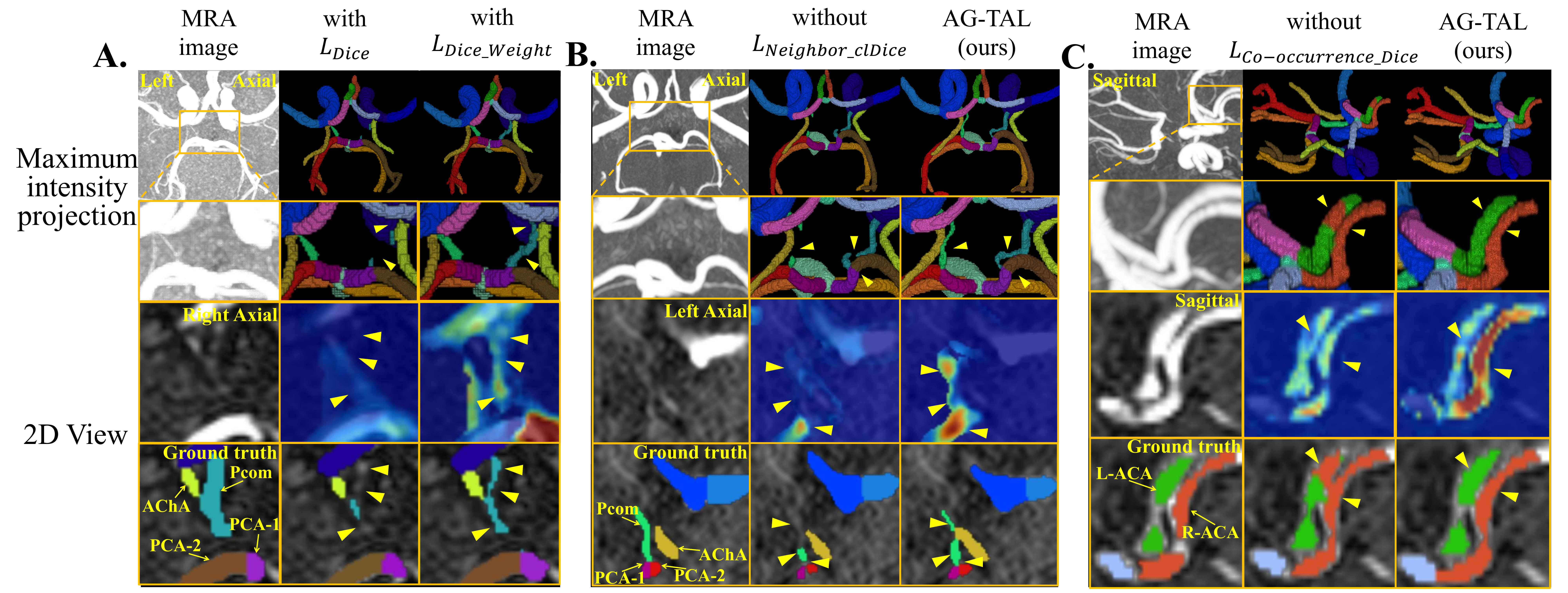}
	\caption{Visualization samples of ablation models. The yellow arrows indicate vascular prone to fragmentation or misclassification. \textbf{(A).} After weighting the Dice loss with vascular radius, increased attention is observed for small vessels with low contrast. \textbf{(B).} Removal of the breakage-aware clDice loss leads to reduced attention to fragile vascular regions and degrades segmentation performance for small vessels. \textbf{(C).} Eliminating the adjacency-aware co-occurrence loss results in uniform attention distribution between easily confusable bilateral arteries, preventing the model from learning inter-class distinctions and introducing misclassification.}
	\label{fig_6}
\end{figure*}

	Grad-CAM visualizations (Fig. \ref{fig_6}) reveal that each loss component induces anatomically meaningful attention consistent with quantitative improvements. The radius-aware Dice loss effectively mitigates class imbalance by shifting attention toward small vessels, improving segmentation continuity in thin regions like the Pcom compared to standard Dice loss (Fig. \ref{fig_6}A). Incorporating breakage-aware clDice further amplifies focus on fragile, low-contrast arteries, and its removal leads to reduced Pcom attention and evident vessel breakage (Fig. \ref{fig_6}B). Finally, the adjacency-aware co-occurrence loss enhances spatial discrimination. Unlike the symmetric responses seen without it, the complete AG-TAL produces localized, asymmetric attention across neighboring arteries, reducing misclassification (Fig. \ref{fig_6}C).

	\subsection{Reliability and robustness analysis}
	SNR analysis on the test set (Fig. \ref{fig_app}A) showed slopes near zero for all vessel scales, indicating minimal sensitivity to image quality. Reliability analysis of average diameter (Fig. \ref{fig_app}B) revealed good-to-excellent ICCs (0.72–0.90+)~\cite{b30} for most arteries across four scans, with an average ICC of 0.93 for the overall CoW diameter.
	The t-SNE visualization demonstrated consistent clustering for the four scans of most subjects. Overall, the SNR and test-retest analysis demonstrate the strong robustness and reliability of AG-TAL.
	
	\subsection{AG-TAL application to AD analysis}
\begin{figure*}
	\includegraphics[width=\textwidth]{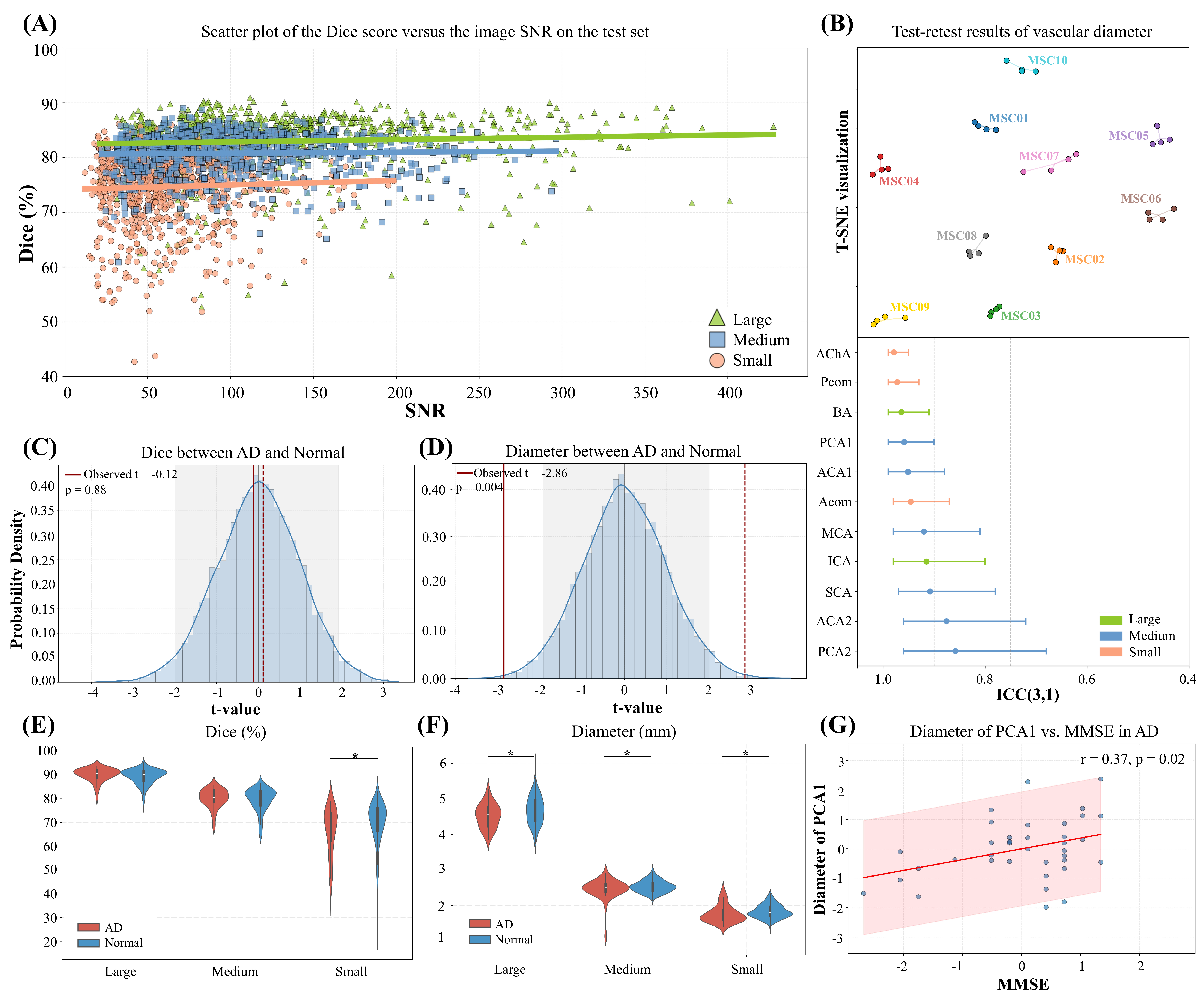}
	\caption{Comprehensive evaluation and application of the AG-TAL. \textbf{(A).} Scatter plot of the Dice versus the SNR for large, medium, and small arteries. The regression lines for all three groups show near-zero slopes. \textbf{(B).} Evaluation of segmentation reproducibility. Forest plot illustrating the test-retest reliability of artery diameter measurements grouped by size (large, medium, small) using ICC(3,1). T-SNE visualizations of the overall diameter across all 10 subjects (MSC01–10) in the MSC dataset. \textbf{(C-D).} Permutation test (10000 permutations) t-value distributions for the Dice metric and mean vascular diameter differences between AD and NC groups. \textbf{(E-F).} Violin plots of the Dice and mean vascular diameter for large, medium, and small arteries. $ *, p < 0.05$ using t test with FDR correction for multiple comparisons. \textbf{(G).} Correlations between PCA1 mean diameter and the MMSE score in AD patients. The scatter plot includes a linear regression line with a shaded area representing the 95\% confidence interval.}
	\label{fig_app}
\end{figure*}

	Comparative analysis between AD and NC cohorts confirmed that segmentation performance was unbiased. Permutation tests on Dice scores (Fig. \ref{fig_app}C) sevealed no significant inter-group differences (observed t = -0.14, p = 0.88), with distributions remaining nearly identical across large and medium arteries (Fig. \ref{fig_app}E) While small arteries showed a minor significant difference (p $  <  $ 0.05) , overall accuracy remained comparable.
	To evaluate morphometric biomarkers, comparisons were restricted to individuals aged $\ge$ 66 to minimize confounding effects. Cohorts were balanced for age (AD: 73.4 $  \pm  $ 6.9; NC: 72.0 $  \pm  $ 4.6 years; p = 0.11) and sex (AD: 24 males / 14 females; NC: 98 males / 102 females; p = 0.15) , with both factors treated as covariates in subsequent analyses. Permutation tests (Fig. \ref{fig_app}D, F) indicated a significant reduction in mean arterial diameter in the AD group (observed t = -2.86, p = 0.004), particularly within the left Pcom (p = 0.02) and bilateral AChA (R-AChA: p = 0.003, L-AChA: p = 0.04).
	Furthermore, vascular morphology correlated with cognitive function in AD patients. Correlation analyses were performed between the average diameters of each vascular segment within the CoW and cognitive assessment scores. A significant positive correlation was identified between mini-mental state examination (MMSE) scores and the average diameter of the PCA1 segment (Fig. \ref{fig_app}G, p = 0.02). These results demonstrate that the AG-TAL provides unbiased segmentation and facilitates the extraction of clinically relevant biomarkers associated with cognitive impairment in AD.

\section{Discussion}

In this study, we propose AG-TAL to address the key challenges in multiclass CoW segmentation: vascular discontinuity, inter-class misclassification, and the performance imbalance of small arteries. 
The superior performance of AG-TAL stems from its multi-level approach, which jointly combines voxel-wise supervision, local-neighborhood consistency, and global anatomical priors to guide the model toward anatomically plausible and topologically continuous segmentations.

A practical advantage of AG-TAL is its implementation as a lightweight, plug-and-play objective~\cite{b35,b36}, requiring no architectural modifications or heavy post-processing. Consequently, AG-TAL is compatible with diverse backbones, including CNNs, Transformer-based or Mamba-based networks, without altering inference pipelines~\cite{b6,b33,b34,b38}. This design ensures clinical deployment efficiency~\cite{b39,b40} and facilitates extension to other imaging modalities and datasets.

AG-TAL incorporates three key components to optimize multiclass CoW segmentation. First, the radius-aware Dice loss mitigates class imbalance by up-weighting voxels of small vessels, reallocating sufficient learning capacity to these vessels. Unlike other radius-driven losses like cbDice, this weighting is pre-computed, significantly reducing training overhead. Second, complementing voxel-level reweighting, the breakage-aware clDice loss adapts efficiently topology-sensitive supervision to multiclass setting. By extracting a shared foreground skeleton and modulating it with class-specific neighborhood inconsistency maps, it enforces continuity constraints without the memory intensity of per-class skeletonization. 
Finally, beyond continuity, the adjacency-aware co-occurrence loss addresses anatomically implausible label assignments by explicitly leveraging neuroanatomical priors via a predefined adjacency matrix, which aligns with prior findings that anatomical priors can reduce false positives in segmentation~\cite{b41,b42,b43}. Importantly, its differentiable co-occurrence feature operates on soft probability maps, enhancing training stability. Moreover, its reliance on anatomical priors makes it naturally extensible to weakly supervised settings where dense annotations are unavailable. Collectively, these components enable AG-TAL to ensure anatomical plausibility and topological continuity with high computational efficiency.

An additional challenge in CoW segmentation is the prevalence of anatomical variants, commonly observed in healthy populations~\cite{b25,b46}. In our cohort, variations were mainly found in Pcom (24.5\%), followed by AChA (3.7\%) and Acom (2.9\%). Overly rigid constraints can induce false positives or incorrect connections. AG-TAL addresses this by balancing local continuity enforcement and global adjacency guidance. The breakage-aware clDice loss suppresses isolated or locally discontinuous false positives, consistent with prior studies~\cite{b9,b49}. Simultaneously, the adjacency-aware co-occurrence loss leverages anatomical priors and ground truth supervision to avoid enforcing incorrect connections for absent arteries, ensuring robustness under vascular variation.

Clinical analyses further support the value of AG-TAL in a neurodegenerative population. AG-TAL achieves robust and unbiased CoW segmentation in AD patients, enabling reliable  downstream morphometric analysis. Segmentations showed significant arterial diameter reductions in AD, particularly in small vessels, consistent with prior studies reporting microvascular impairment~\cite{b46,b50} and vessel narrowing~\cite{b46,b52}. Reduced PCA1 diameter was found to be positively correlated with lower MMSE, consistent with posterior circulation hypoperfusion and posterior cortical involvement in AD~\cite{b53,b54}. Since PCA1 perfuses posterior cortical regions involved in global cognition~\cite{b54}, these associations suggest that vascular morphological alterations may reflect or contribute to region-specific dysfunction. Overall, AG-TAL facilitates extraction of clinically meaningful vascular biomarkers and may support imaging-based disease characterization.

Beyond methodological contributions, this work also introduces a multi-center, manually annotated CoW dataset across multiple institutions under a unified protocol. By addressing the lack of detailed CoW annotations~\cite{b20,b21,b23} and diversity of imaging centers~\cite{b25} in existing public MRA datasets, it provides a robust foundation for evaluating model generalizability under heterogeneous conditions.

Several limitations should be acknowledged. First, the segmentation consistency of AG-TAL is affected by variability in the annotation of open-ended arteries (e.g., SCA, PCA2, and ACA2). Due to anatomical heterogeneity and annotation subjectivity, endpoint definitions are not uniform across subjects, preventing the model from learning consistent cutoff patterns. Future work will address this issue by refining annotation protocols to enforce stricter anatomical boundary definitions. Second, the adjacency-aware co-occurrence loss relies on a predefined connectivity matrix based on standard neuroanatomy. While such anatomical priors are central to AG-TAL’s performance, they inherently limit generalizability to scenarios where foreground tubular connectivity is well characterized, and may not extend to structures with unknown or ambiguous spatial relationships.

In conclusion, AG-TAL provides a robust and effective solution for multiclass CoW segmentation, enhancing topological integrity and anatomical rationality when combined with our multi-center annotated dataset. By enabling precise, artery-specific morphometry, AG-TAL facilitates the characterization of vascular variations and supporting downstream analysis in cerebrovascular disease research.


\section*{References}

\begin{thebibliography}{00}
	\bibitem{b1} K. Malhotra, J. Gornbein, and J. L. Saver, ``Ischemic Strokes Due to Large-Vessel Occlusions Contribute Disproportionately to Stroke-Related Dependence and Death: A Review'', \emph{Front. Neurol.}, vol. 8, pp. 651, Nov. 2017.
	
	\bibitem{b2} M. J. R. Lim, C. S. Tan, B. Gyanwali, C. Chen, and S. Hilal, ``The effect of intracranial stenosis on cognitive decline in a memory clinic cohort'', \emph{Eur. J. Neurol.}, vol. 28, no. 6, pp. 1829–1839, Jun. 2021.
	
	\bibitem{b3} V. L. Gray~\emph{et al.}, ``Asymptomatic carotid stenosis is associated with mobility and cognitive dysfunction and heightens falls in older adults'', \emph{J. Vasc. Surg.}, vol. 71, no. 6, pp. 1930–1937, Jun. 2020.
	
	
	\bibitem{b4} F. Dumais~\emph{et al.}, ``eICAB: A novel deep learning pipeline for Circle of Willis multiclass segmentation and analysis'', \emph{NeuroImage}, vol. 260, pp. 119425, Oct. 2022.
	
	\bibitem{b5} J. Hou~\emph{et al.}, ``Relationship between Circle of Willis Variations and Cerebral or Cervical Arteries Stenosis Investigated by Computer Tomography Angiography and Multitask Convolutional Neural Network'', \emph{J. Healthc. Eng.}, vol. 2021, pp. 1–8, Oct. 2021.
	
	
	\bibitem{b6} F. Isensee, P. F. Jaeger, S. A. A. Kohl, J. Petersen, and K. H. Maier-Hein, ``nnU-Net: a self-configuring method for deep learning-based biomedical image segmentation'', \emph{Nat. methods}, vol. 18, no. 2, pp. 203–211, Feb. 2021.
	
	
	\bibitem{b7} A. H. Berger~\emph{et al.}, ``Topologically faithful multi-class segmentation in medical images,'' in \emph{Proc. Int. Conf. Med. Image Comput. Comput.-Assist. Intervent. (MICCAI)}, Oct. 2024, pp. 721–731.
	
	
	\bibitem{b8} J. R. Clough, N. Byrne, I. Oksuz, V. A. Zimmer, J. A. Schnabel, and A. P. King, ``A Topological Loss Function for Deep-Learning Based Image Segmentation Using Persistent Homology'', \emph{IEEE Trans. Pattern Anal. Mach. Intell.}, vol. 44, no. 12, pp. 8766–8778, Dec. 2022.
	
	
	\bibitem{b9} X. Hu, F. Li, D. Samaras, and C. Chen, ``Topology-preserving deep image segmentation,'' in \emph{Advances in Neural Information Processing Systems (NeurIPS)}, Vancouver, BC, Canada, Dec. 2019.
	
	\bibitem{b10} N. Stucki, J. C. Paetzold, S. Shit, B. H. Menze, and U. Bauer, ``Topologically faithful image segmentation via induced matching of persistence barcodes'', in \emph{Proc. 40th Int. Conf. Mach. Learn. (ICML)}, Jul. 2023, pp. 32698–32727.
	
	
	\bibitem{b11} N. Stucki, V. Bürgin, J. C. Paetzold, and U. Bauer, ``Efficient betti matching enables topology-aware 3D segmentation via persistent homology'', \emph{arXiv preprint arXiv:2407.04683}, 2024.
	
	\bibitem{b12} S. Shit~\emph{et al.}, ``clDice - a Novel Topology-Preserving Loss Function for Tubular Structure Segmentation'', in \emph{Proc. IEEE/CVF Conf. Comput. Vis. Pattern Recognit. (CVPR)}, Nashville, TN, USA, Jun. 2021, pp. 16555–16564.
	
	\bibitem{b13} M. J. Menten~\emph{et al.}, ``A skeletonization algorithm for gradient-based optimization,'' in \emph{Proc. IEEE/CVF Int. Conf. Comput. Vis. (ICCV)}, 2023, pp. 21394–21403.
	
	\bibitem{b14} P. Shi~\emph{et al.}, ``Centerline boundary dice loss for vascular segmentation,'' in \emph{Int. Conf. Med. Image Comput. Comput.-Assist. Intervent. (MICCAI)}, Cham, Switzerland: Springer Nature Switzerland, 2024, pp. 46–56.
	
	\bibitem{b15} C. Acebes, A. H. Moustafa, O. Camara, and A. Galdran, ``The centerline-cross entropy loss for vessel-like structure segmentation: Better topology consistency without sacrificing accuracy,'' in \emph{Int. Conf. Med. Image Comput. Comput.-Assist. Intervent. (MICCAI)}, Cham, Switzerland: Springer Nature Switzerland, 2024, pp. 710–720.
	
	\bibitem{b16} Y. Kirchhoff~\emph{et al.}, ``Skeleton recall loss for connectivity conserving and resource efficient segmentation of thin tubular structures,'' in \emph{Proc. Eur. Conf. Comput. Vis. (ECCV)}, Cham, Switzerland: Springer Nature Switzerland, 2024, pp. 218–234.
	
	
	\bibitem{b17} H. Edelsbrunner, \emph{Computational topology: an introduction}. in Miscellaneous Books, vol. 69. Providence, R.I: American Mathematical Society, 2010.
	
	\bibitem{b24} UMC Utrecht, \emph{Aneurysm Detection And segMentation Challenge (ADAM) 2020}, [Online]. Available: https://adam.isi.uu.nl/. Accessed: Oct. 16, 2025.
	
	\bibitem{b18} S. N. Wright~\emph{et al.}, ``Digital reconstruction and morphometric analysis of human brain arterial vasculature from magnetic resonance angiography'', \emph{NeuroImage}, vol. 82, pp. 170–181, Nov. 2013.
	
	\bibitem{b19} T. Di Noto~\emph{et al.}, ``Towards Automated Brain Aneurysm Detection in TOF-MRA: Open Data, Weak Labels, and Anatomical Knowledge'', Neuroinformatics, vol. 21, no. 1, pp. 21–34, Jan. 2023.
	
	\bibitem{b20} Mazziotta~\emph{et al.}, ``A probabilistic atlas and reference system for the human brain: International Consortium for Brain Mapping (ICBM),'' \emph{Philos. Trans. R. Soc. Lond. B, Biol. Sci.}, vol. 356, pp. 1293–1322, 2001.
	
	\bibitem{b21} Imperial College London, \emph{IXI Dataset – Brain Development}, [Online]. Available: https://brain-development.org/ixi-dataset/. Accessed: Oct. 16, 2025.
	
	\bibitem{b22} E. Bullitt~\emph{et al.}, ``The effects of healthy aging on intracerebral blood vessels visualized by magnetic resonance angiography,'' \emph{Neurobiol. Aging}, vol. 31, no. 2, pp. 290–300, Feb. 2010.
	
	\bibitem{b23} P. J. LaMontagne~\emph{et al.}, ``OASIS-3: Longitudinal neuroimaging, clinical, and cognitive dataset for normal aging and Alzheimer disease,'' \emph{Radiology and Imaging}, Dec. 15, 2019.
	
	
	\bibitem{b25} K. Yang \emph{et al.}, ``Benchmarking the CoW with the TopCoW Challenge: Topology-aware anatomical segmentation of the Circle of Willis for CTA and MRA,'' \emph{arXiv preprint arXiv:2312.17670}, 2023.
	
	\bibitem{b29} E. M. Gordon~\emph{et al.}, ``Precision functional mapping of individual human brains,'' \emph{Neuron}, vol. 95, no. 4, pp. 791–807, 2017.
	
	\bibitem{b26} B. B. Avants,~\emph{ANTs Wiki}, GitHub repository. [Online]. Available: https://github.com/ANTsX/ANTs/wiki. Accessed: Oct. 16, 2025.
	
	\bibitem{b27} J. Meng~\emph{et al.}, ``Highly accurate, automated quantification of 2D/3D orientation for cerebrovasculature using window optimizing method,'' \emph{J. Biomed. Opt.}, vol. 27, 2022.
	
	\bibitem{b28} H. Zhang, H. Zhang, C. Wang, and J. Xie, ``Co-occurrent features in semantic segmentation,'' in \emph{Proc. IEEE/CVF Conf. Comput. Vis. Pattern Recognit. (CVPR)}, 2019, pp. 548–557.
	
	\bibitem{b33} A. Hatamizadeh~\emph{et al.}, ``Swin UNETR: Swin transformers for semantic segmentation of brain tumors in MRI images,'' in \emph{Int. MICCAI Brainlesion Workshop}, Cham, Switzerland: Springer International Publishing, 2021, pp. 272–284.
	
	\bibitem{b34} J. Ma, F. Li, and B. Wang, ``U-Mamba: Enhancing long-range dependency for biomedical image segmentation,'' \emph{arXiv preprint arXiv:2401.04722}, 2024.
	
	\bibitem{b32} R. R. Selvaraju~\emph{et al.}, ``Grad-CAM: Visual explanations from deep networks via gradient-based localization,'' in \emph{Proc. IEEE Int. Conf. Comput. Vis. (ICCV)}, 2017, pp. 618–626.
	
	\bibitem{b30} T. K. Koo and M. Y. Li, ``A guideline of selecting and reporting intraclass correlation coefficients for reliability research,'' \emph{J. Chiropr. Med.}, vol. 15, no. 2, pp. 155–163, Jun. 2016.
	
	
	\bibitem{bffcm} Y. Cui~\emph{et al.}, ``FFCM-MRF: An accurate and generalizable cerebrovascular segmentation pipeline for humans and rhesus monkeys based on TOF-MRA,'' \emph{Comput. Biol. Med.}, vol. 170, p. 107996, 2024.
	
	\bibitem{b35} J. Ma~\emph{et al.}, ``Loss odyssey in medical image segmentation,'' \emph{Med. Image Anal.}, vol. 71, p. 102035, Jul. 2021.
	
	\bibitem{b36} R. Azad~\emph{et al.}, ``Loss functions in the era of semantic segmentation: A survey and outlook,'' \emph{arXiv preprint arXiv:2312.05391}, 2023.
	
	
	\bibitem{b38} R. Azad~\emph{et al.}, ``Medical image segmentation review: The success of U-Net,'' \emph{IEEE Trans. Pattern Anal. Mach. Intell.}, 2024.
	
	\bibitem{b39} D. Karimi and S. E. Salcudean, ``Reducing the Hausdorff distance in medical image segmentation with convolutional neural networks,'' \emph{IEEE Trans. Med. Imaging}, vol. 39, no. 2, pp. 499–513, Feb. 2020.
	
	\bibitem{b40} H. Kervadec~\emph{et al.}, ``Boundary loss for highly unbalanced segmentation,'' \emph{Med. Image Anal.}, vol. 67, p. 101851, Jan. 2021.
	
	\bibitem{b41} T. S. Mathai, B. Liu, and R. M. Summers, ``Segmentation of mediastinal lymph nodes in CT with anatomical priors,'' \emph{Int. J. Comput. Assist. Radiol. Surg.}, vol. 19, no. 8, pp. 1537–1544, May 2024.
	
	\bibitem{b42} P.-A. Ganaye, M. Sdika, B. Triggs, and H. Benoit-Cattin, “Removing segmentation inconsistencies with semi-supervised non-adjacency constraint,” Med. Image Anal., vol. 58, Art. no. 101551, 2019.
	
	\bibitem{b43} H. Zhao, A. Wang, and C. Zhang, ``Research on melanoma image segmentation by incorporating medical prior knowledge,'' \emph{PeerJ Comput. Sci.}, vol. 8, p. e1122, Oct. 2022.
	
	\bibitem{b46} A. Deshpande~\emph{et al.}, ``Novel imaging markers for altered cerebrovascular morphology in aging, stroke, and Alzheimer’s disease,'' \emph{J. Neuroimaging}, vol. 32, no. 5, pp. 956–967, Sep. 2022.
	
	
	
	\bibitem{b49} Y. Zhou \emph{et al.}, ``CF-loss: Clinically-relevant feature optimised loss function for retinal multi-class vessel segmentation and vascular feature measurement,'' \emph{Med. Image Anal.}, vol. 93, p. 103098, Apr. 2024.
	
	\bibitem{b50} T. Kim \emph{et al.}, ``Quantitative analysis of cerebral vasculature and its clinical effect on cognitive change in Alzheimer’s dementia and mild cognitive impairment,'' \emph{J. Neurol. Sci.}, vol. 476, p. 123598, Sep. 2025.
	
	\bibitem{b52} K. M. Kang \emph{et al.}, ``Intracranial stenosis and longitudinal progression of Alzheimer’s disease pathologies,'' \emph{Alzheimers Dement.}, vol. 21, no. 12, p. e71010, Dec. 2025.
	
	\bibitem{b53} D. C. Alsop, W. Dai, M. Grossman, and J. A. Detre, ``Arterial spin labeling blood flow MRI: Its role in the early characterization of Alzheimer’s disease,'' \emph{J. Alzheimers Dis.}, vol. 20, no. 3, pp. 871–880, May 2010.
	
	\bibitem{b54} D. C. Alsop, J. A. Detre, and M. Grossman, ``Assessment of cerebral blood flow in Alzheimer’s disease by spin-labeled magnetic resonance imaging,'' \emph{Ann. Neurol.}, vol. 47, no. 1, pp. 93–100, Jan. 2000.
	
	
\end{thebibliography}
\end{document}